\documentclass[floatfix,aps,prl,twocolumn,superscriptaddress,a4paper,english]{revtex4}

\usepackage{times}
\usepackage{color}
\usepackage[colorlinks=true,urlcolor=blue,citecolor=blue]{hyperref}
\usepackage{url}
\usepackage{breakurl}
\usepackage{soul}
\usepackage{graphicx}
\usepackage{amsmath}
\usepackage{verbatim}
\usepackage{subfigure}
\usepackage{xcolor}
\usepackage{amssymb}
\usepackage{latexsym}
\usepackage{hyperref}
\DeclareGraphicsExtensions{.png,.eps}
\usepackage{float}

\usepackage{xcolor}
\usepackage[urlcolor=blue]{hyperref}      
\hypersetup{
    colorlinks = true,                    
    citecolor = {blue},
    linkcolor = {purple},
           }
\begin{document}	
	
\title{Tangent-Space Gradient Optimization of Tensor Network for Machine Learning}

\author{Zheng-Zhi Sun}
\affiliation{School of Physical Sciences, University of Chinese Academy of Sciences, P. O. Box 4588, Beijing 100049, China}

\author{Shi-Ju Ran}\email[Corresponding Author. Email: ]{sjran@cnu.edu.cn}
\affiliation{Department of Physics, Capital Normal University, Beijing 100048, China}

\author{Gang Su}
\email[Corresponding author. Email: ] {gsu@ucas.ac.cn}
\affiliation{School of Physical Sciences, University of Chinese Academy of Sciences, P. O. Box 4588, Beijing 100049, China}
\affiliation{Kavli Institute for Theoretical Sciences, and CAS Center for Excellence in Topological Quantum Computation, University of Chinese Academy of Sciences, Beijing 100190, China}

\date{\today}

\begin{abstract}

     The gradient-based optimization method for {{deep machine learning models}} suffers from gradient vanishing and exploding problems, particularly when the computational graph becomes deep. In this work, we propose the tangent-space gradient optimization (TSGO) for the probabilistic models to keep the gradients from vanishing or exploding. The central idea is to guarantee the orthogonality between the variational parameters and the gradients. The optimization is then implemented by rotating parameter vector towards the direction of gradient. We explain and testify TSGO in tensor network (TN) machine learning, where the TN describes the joint probability distribution as a normalized state $\left| \psi  \right\rangle $ in Hilbert space. We show that the gradient can be restricted in the tangent space of $\left\langle \psi  \right.\left| \psi  \right\rangle  = 1$ hyper-sphere.  Instead of additional adaptive methods to control the learning rate in deep learning, the learning rate of TSGO is naturally determined by the angle $\theta $ as $\eta  = \tan \theta $. Our numerical results reveal better convergence of TSGO in comparison to the off-the-shelf Adam.

\end{abstract}

\maketitle

\textit{Introduction.---}The gradient based optimization is of fundamental importance to many fields of science and engineering \cite{lecun2015deep,SIG-039,10.5555/3104482.3104516,QIAN1999145,KIVINEN19971,burges2005learning,10.1007/978-3-7908-2604-3_16}. In particular, the back-propagation (BP) algorithm is widely used in training feedforward neural networks \cite{Rojas1996,Goodfellow-et-al-2016,HECHTNIELSEN199265}, which are applied to many fields from computer vision to board game programs and achieve competitive or superior results compared with human experts \cite{DBLP:conf/cvpr/CiresanMS12, NIPS2012_4824,doi:10.5858/arpa.2016-0471-ED}. However, BP algorithm suffers from the well-known gradient vanishing and exploding problems, particularly when the computational graph becomes deep \cite{Goodfellow-et-al-2016}, which makes the optimization inefficient or unstable. Therefore, the stochastic gradient-based optimization methods to properly determine the learning rate, such as stochastic gradient descent \cite{10.2307/2236626,kushner2003stochastic}, root mean square propagation \cite{tieleman2012lecture}, adaptive learning rate method \cite{zeiler2012adadelta}, and adaptive moment estimation (Adam) \cite{DBLP:journals/corr/KingmaB14}, are proposed to keep the gradients from vanishing and exploding. Still, the validity of these methods including Adam still depends on the manual choices of the learning rate \cite{Goodfellow-et-al-2016}.

Tensor network (TN), which is a powerful numerical tool for quantum many-body physics and quantum information sciences \cite{verstraete2008matrix, ran2017review, Evenbly2011, bridgeman2017hand, SCHOLLWOCK201196,Cirac_2009, ORUS2014117}, has been recently applied to machine learning \cite{MAL-059,MAL-067,biamonte2017quantum,Huggins_2019,NIPS2016_6211,glasser2018supervised,PhysRevB.97.085104,stoudenmire2018learning,PhysRevX.8.031012,liu2018learning,PhysRevE.98.042114,Liu_2019,PhysRevB.99.155131,pestun2017tensor}. One critical issue under hot debate is the possible advantages of TN over machine learning methods such as gradient-based neural networks \cite{Goodfellow-et-al-2016, goodfellow2013empirical}. For the unsupervised learning as an example, TN uses a different strategy from neural network (NN), e.g., the generative adversarial networks \cite{NIPS2014_5423} or pixel convolutional NN's \cite{TANG2018125}, which is explicitly modeling the joint probability distribution of the features as a quantum many-body state or ``Born machine'' \cite{e20080583, PhysRevB.97.085104, PhysRevB.99.155131, PhysRevX.8.031012}. In this way, the statistical properties including correlations and entropies can be readily extracted from the TN \cite{Evenbly2011}, This, in general, cannot be done with NN as it represents a complicated non-linear map.

\begin{figure}[htb]
	\includegraphics[width=0.5\linewidth]{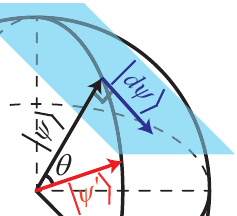}
	\caption{\label{instance-new} A sketch of updating (rotating) the state $|\psi \rangle$ to $|\psi' \rangle$ with an angle $\theta $. The rotation direction $|d\psi\rangle$ is the gradient direction which is orthogonal to $|\psi \rangle$. Its proof is given in text.}
\end{figure}

In this work, we introduce the tangent-space gradient optimization (TSGO) as a gradient-based method for probabilistic models. The TSGO optimizes a parameter vector by rotating it towards the direction of gradient, which is guaranteed to be in the tangent hyperplane of the parameter space. The learning rate $\eta $ is then controlled by the rotation angle $\theta$ through $\eta = \tan \theta$. This in general avoids the gradient vanishing or exploding problems and promises a robust way to determine the learning rate. For the TN generative model \cite{PhysRevX.8.031012, PhysRevB.99.155131}, the probability distribution is described by a normalized state (denoted as $|\psi \rangle$) in Hilbert space. The normalization of the state $\langle \psi |\psi \rangle= 1$ (i.e., the normalization of the probability distribution) can be easily done using the central-orthogonal form of the TN \cite{DBLP:journals/qic/Perez-GarciaVWC07, PhysRevA.74.022320, PhysRevLett.100.240603}. Then the gradient is proved to be on the tangent hyperplane of the sphere satisfying $\langle \psi |\psi \rangle= 1$. The optimization process is shown in Fig. \ref{instance-new}.

\textit{Preliminaries.---} Denoting the variational parameters of the probabilistic model to be updated as $W$ (written as a vector or a tensor), we propose the TSGO by which the gradients satisfy
\begin{eqnarray}
\label{eq-TSGO}
\langle W, {{\partial f} \over {\partial W}} \rangle = 0,
\end{eqnarray}
where $f$ is the loss function and $\langle *, * \rangle $ means the inner product of two vectors or two tensors with summing over all indexes correspondingly. In other terms, TSGO requires that the gradients are orthogonal to the parameter vector.

Before demonstrating how the orthogonality avoids the gradient vanishing and exploding problems, let us first discuss the conditions that satisfy Eq. (\ref{eq-TSGO}). We consider $f$ as a functional of the probability distribution of the samples, which can be formally written as
\begin{eqnarray}
\label{eq-functional}
f = \sum\limits_{X \in {\cal A}} {F\left[ {P(X;W)} \right]} ,
\end{eqnarray}
with $X$ the samples in the training set $\mathcal{A}$ which contains $A$ samples. Then we denote that a sufficient condition for Eq. (\ref{eq-TSGO}) can be written as
\begin{eqnarray}
\label{eq-condition}
P\left( {X;W} \right) = P\left( {X;\alpha W} \right),
\end{eqnarray}
for any sample $X$ and any non-zero constant $\alpha$. In other words, the TSGO can be implemented when any nonzero constant scaling of the parameters $W$ does not affect the probability distribution. The proof is given as follows.

The directional derivative of $P\left( {X;W} \right)$ along the parameter vector $W$ can be written as
\begin{eqnarray}
\label{eq-df1}
{\partial _W}P\left( {X;W} \right){\rm{ = }}\mathop {\lim }\limits_{h \to 0} {{P\left( {X;W + hW} \right) - P\left( {X;W} \right)} \over h}.
\end{eqnarray}
When Eq. (\ref{eq-condition}) is satisfied, it can be easily seen that ${\partial _W}P\left( {X;W} \right){\rm{ = 0}}$ since $P(X; W+hW) - P(X; W) = 0$. Now we write the direction derivative in another equivalent form as
\begin{eqnarray}
\label{eq-df2}
{\partial _W}P\left( {X;W} \right){\rm{ = }} \langle W, {{\partial P\left( {X;W} \right)} \over {\partial W}} \rangle.
\end{eqnarray}
Then we have
\begin{eqnarray}
\label{eq-final}
\langle W,{{\partial f} \over {\partial W}}\rangle  = \sum\limits_{X \in {\cal A}} {{{\partial F} \over {\partial P\left( {X;W} \right)}}} \langle W,{{\partial P\left( {X;W} \right)} \over {\partial W}}\rangle  = 0.
\end{eqnarray}
Thus the gradients of a probabilistic model satisfying Eq. (\ref{eq-condition}) are orthogonal to the parameter vector, where TSGO can be implemented.

\textit{TSGO for tensor network machine learning.---} To further explain TSGO, we implement it on the unsupervised TN machine learning, where the TN is used to capture the joint probability distribution of features in Hilbert space \cite{PhysRevX.8.031012, PhysRevB.99.155131}. The orthogonal form of the TN can be utilized to satisfy Eq. (\ref{eq-condition}).

Let us start with some necessary preliminaries of unsupervised TN machine learning methods. The first step for TN machine learning is to map the data onto the Hilbert space. We take images as an example. One feature (pixel of images) $x \in [0, 1]$ is mapped to the state of a qubit, i.e., $x \to |x\rangle = \cos (x \pi /2) |0\rangle + \sin (x \pi /2) |1\rangle $, with $|0\rangle$ and  $|1\rangle$ the eigenstates of the Pauli matrix $\hat{\sigma}^z$ \cite{NIPS2016_6211}. In this way, one image is mapped to a product state $|X\rangle = \prod_{\otimes n} |x_n\rangle$, with $x_n$ the $n$-th pixel of the image.

For a specific task of, e.g., generating images of hand-written digits, TN machine learning aims to model the joint probability distribution $P(x_1, \cdots, x_N)$ of the features (with $N$ the number of features). The strategy is to represent $P$ with a many-body state $|\psi\rangle $ \cite{PhysRevX.8.031012}. The probability of a given sample $X = (x_1, \cdots, x_N)$ is represented with the square of the amplitude
\begin{eqnarray}
\label{eq-px}
P(X) = \frac{\langle X|\psi\rangle^2}{\langle \psi | \psi \rangle},
\end{eqnarray}
in accordance to Born's probabilistic interpretation of quantum wave-functions \cite{born1926quantum, RevModPhys.42.358}.

With a given set of samples, $|\psi\rangle$ is optimized by minimizing a loss function that describes the difference between the joint probability distribution from the training set $\mathcal{A}$ and $P$. One common choice of loss function is the negative-log likelihood (NLL) \cite{10.2307/2236703}
\begin{eqnarray}
\label{eq-NLL}
f =  - {1 \over A}\sum\limits_{X \in {\cal A}} {\log } P(X).
\end{eqnarray}

To efficiently represent and update $|\psi\rangle$, the coefficients are written in a compact form of TN. We here choose the matrix product state (MPS) \cite{DBLP:journals/qic/Perez-GarciaVWC07,doi:10.1137/090752286} as an example to represent the many-body state. The coefficients of TN in MPS form can be written as follows
\begin{eqnarray}
\label{eq-MPS}
{\psi _{{s_1}{s_2} \cdots {s_N}}} = \sum\limits_{{\alpha _0},{\alpha _1}, \cdots ,{\alpha _N}} {T_{{\alpha _0}{s_1}{\alpha _1}}^{[1]}} T_{{\alpha _1}{s_2}{\alpha _2}}^{[2]} \cdots T_{{\alpha _{N - 1}}{s_N}{\alpha _N}}^{[N]}.
\end{eqnarray}
The optimization of $\left| \psi  \right\rangle $ becomes the optimization of tensors $\{T^{[n]}\}$.

To remove the redundant degrees of freedom from MPS form of a many-body state, we transfer MPS to its canonical form with a gauge transformation \cite{DBLP:journals/qic/Perez-GarciaVWC07,ran2017review}. Then the tensors satisfy the following orthogonal conditions
\begin{eqnarray}
\label{eq-ortLR}
\sum\limits_{{\alpha _{n - 1}}{s_n}} {T_{{\alpha _{n - 1}}{s_n}{\alpha _n}}^{[n]}} T_{{\alpha _{n - 1}}{s_n}{\alpha _{n'}}}^{[n] * } = {\delta _{{\alpha _n}{\alpha _{n'}}}}{\rm{   for }}(n < \tilde n), \\
\sum\limits_{{s_n}{\alpha _n}} {T_{{\alpha _{n - 1}}{s_n}{\alpha _n}}^{[n]}} T_{{\alpha _{n - 1}}{s_n}{\alpha _{n'}}}^{[n] * } = {\delta _{{\alpha _{n - 1}}{\alpha _{n' - 1}}}}{\rm{   for }}(n > \tilde n),
\end{eqnarray}
with $\tilde{n}$ the orthogonal center. The norm of $|\psi\rangle$ becomes the norm of the orthogonal central tensor, i.e., $\langle \psi |\psi \rangle  = \langle {T^{[\tilde n]}},{T^{[\tilde n]}}\rangle = 1$.


We now verify that MPS satisfies the requirement of TSGO. The gradient of loss function Eq. (\ref{eq-NLL}) reads
\begin{eqnarray}
\label{eq-sample}
{{\partial f} \over {\partial |\psi \rangle }} = 2\left| \psi  \right\rangle  - {2 \over A}\sum\limits_{X \in {\cal A}} {{{\left| X \right\rangle } \over {\left\langle X \right.\left| \psi  \right\rangle }}} .
\end{eqnarray}
The gradient satisfies
\begin{eqnarray}
\label{eq-sample}
\left\langle {|\psi \rangle ,{{\partial f} \over {\partial | \psi\rangle }}} \right\rangle  = 2\left\langle \psi  \right.\left| \psi  \right\rangle  - {2 \over A}\sum\limits_{X \in {\cal A}} {{{\left\langle \psi  \right.\left| X \right\rangle } \over {\left\langle X \right.\left| \psi  \right\rangle }}}  = 0.
\end{eqnarray}

In fact, we cannot update the whole MPS, nor calculate the gradient $\frac{\partial f}{\partial |\psi \rangle}$, since the complexity is exponentially high. Luckily, it is easy to show that by only updating the tensor $T^{[\tilde{n}]}$ at the canonical center (the other parameters fixed), one may find that the requirement of TSGO is also satisfied. Using the orthogonal conditions, the probability distribution [Eq. (\ref{eq-px})] becomes $P(X) = \frac{\langle X|\psi\rangle^2}{\langle T^{[\tilde{n}]}, T^{[\tilde{n}]} \rangle}  $. One can show similarly that $\langle {{\partial f} \over {\partial {T^{[\tilde n]}}}},{T^{[\tilde n]}}\rangle  = 0$.

As addressed above, we take MPS as an example to represent $|\psi \rangle$. We stress here that TSGO can be readily implemented on other TN's. The conditions are: (1) the loss function to be minimized is a functional of the probability distribution of the samples; (2) the normalization of $|\psi \rangle$, i.e., $\langle \psi | \psi \rangle$, becomes the norm of one single tensor; (3) any tensor can represent the norm of $|\psi\rangle$ by transforming the TN without errors or with controlled errors. For MPS, we use the central orthogonal form of MPS to satisfy (2), and use gauge transformation to satisfy (3). For a TN without loops, similar orthogonal form can be defined and similar gauge transformations can be done to move the center \cite{PhysRevA.74.022320,PhysRevB.80.235127}, thus (2) and (3) can be satisfied. For loopy TN's such as PEPS \cite{verstraete2004renormalization,PhysRevLett.101.250602,PhysRevB.98.085155,haghshenas2019canonicalization}, recent progresses show that the orthogonal form can still be defined, but the transformations to move the center will inevitably introduce certain numerical errors.

With the TN scheme, we now explain how TSGO avoids the vanishing and exploding of the gradients by a rotational scheme. When $T^{[\tilde{n}]}$ is changed with ${T^{[\tilde n]}} \leftarrow ({T^{[\tilde n]}} - \eta {{\partial f} \over {\partial {T^{[\tilde n]}}}})$, where $\eta $ is the learning rate, the change of the state $\left| {d\psi } \right\rangle $ is in fact in the tangent space of the $\langle \psi | \psi \rangle=1$ hyper-sphere. The reason is that $\langle d\psi |\psi \rangle  =  - \eta \langle {{\partial f} \over {\partial {T^{[\tilde n]}}}},{T^{[\tilde n]}}\rangle  = 0$ by use of the left and right orthogonal conditions.


The optimizations of the central tensor can be re-interpreted as rotations in Hilbert space. The learning rate is controlled by the rotation angle. To rotate $|\psi \rangle$ towards the direction of $\left| {d\psi } \right\rangle $, we update the central tensor $T^{[\tilde{n}]}$ with ${T^{[\tilde n]}} \leftarrow ({T^{[\tilde n]}} - \eta {{\partial f} \over {\partial {T^{[\tilde n]}}}})$. Then we normalize the tensor as $T^{[\tilde{n}]} \leftarrow \frac{T^{[\tilde{n}]}}{|T^{[\tilde{n}]}|}$. From the geometrical relations shown in Fig. \ref{instance-new}, it can be readily seen that the learning rate $\eta$ and rotation angle $\theta$ obey
\begin{eqnarray}
\label{eq-angle}
\eta = \tan \theta.
\end{eqnarray}

The learning rate can be robustly controlled by the rotation angle that is naturally bounded as $0 < \theta  \ll {\pi  \over 2}$ (``$\ll$'' is taken because the learning rate is a small number) . For instance, $\theta$ can be taken as $\pi /6 $ initially. When the loss function increases (meaning $|\psi \rangle$ is over rotated), the rotation angle reduces to its one third. In this way, the change of $|\psi \rangle$ is strictly controlled by $\theta$, and the vanishing and exploding problems of the gradient are avoided. 

To update any tensor in the MPS, we should implement the gauge transformation, which can move the center to any tensor without changing the state $\left| \psi  \right\rangle $. One may refer to Ref. \cite{ran2017review} for details of the gauge transformation. In this way, all tensors in the form of MPS can be optimized with TSGO.

\begin{figure}[htb]
	\includegraphics[width=1\linewidth]{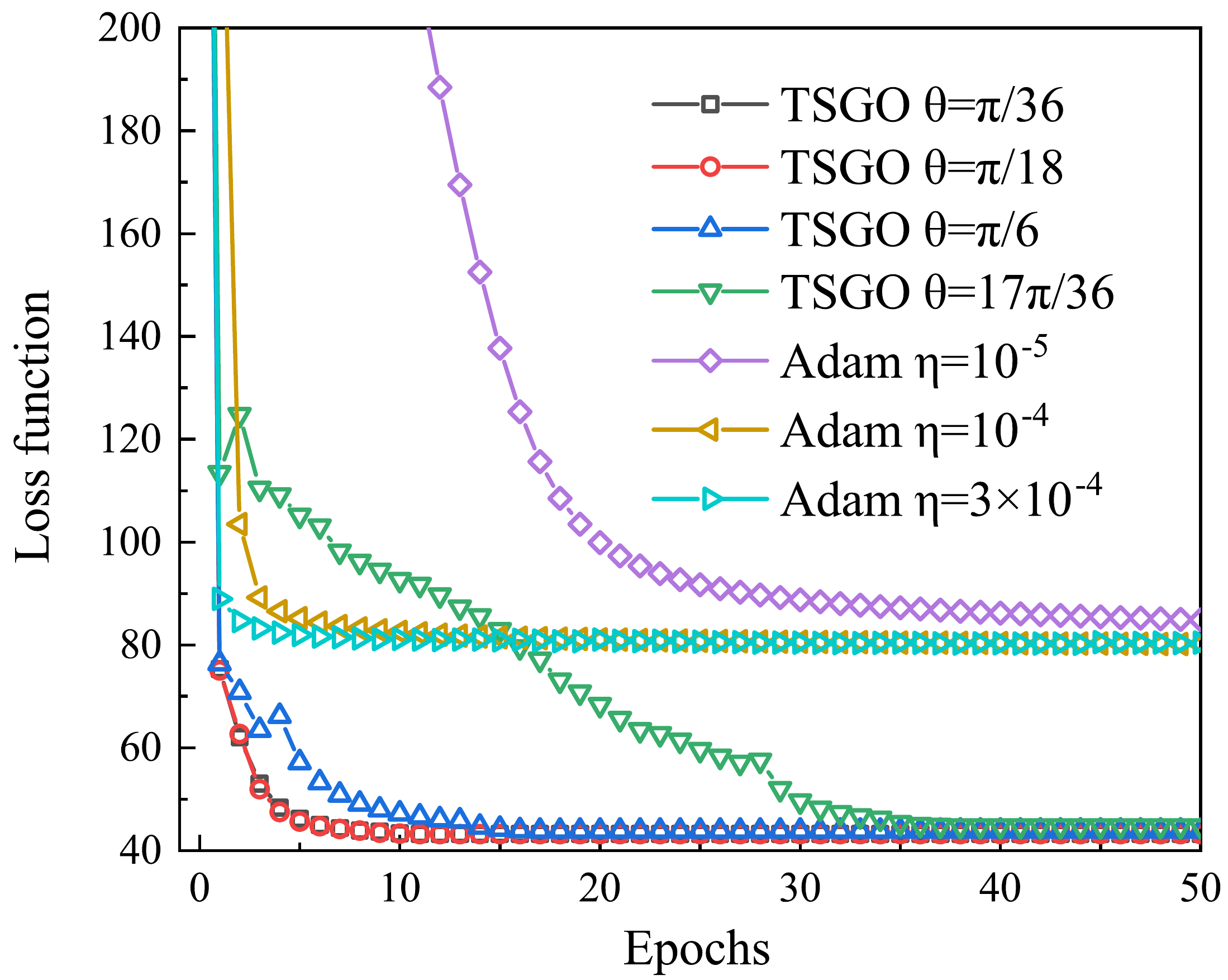}
	\caption{\label{results} Loss function versus epoch by TSGO and Adam with different learning rates (or rotation angles). 6000 images randomly selected from the original MNIST dataset are used in the optimizations. The size of each tensor in the networks is constrained in  $30 \times 2 \times 30$.}
	\end{figure}

\textit{Numerical experiments.---}We compare the convergence between TSGO and BP algorithm \cite{Rojas1996} under the unsupervised generative MPS model \cite{PhysRevX.8.031012} on the MNIST dataset \cite{6296535}. The BP algorithm directly calculate the gradient of the tensors with the auto-gradient method without enforcing the central-orthogonal form of MPS. The learning rate of BP algorithm is determined by Adam \cite{DBLP:journals/corr/KingmaB14}. The numerical results are shown in Fig .\ref{results}.

The TSGO algorithm shows the best convergence in Fig. \ref{results}. TSGO converges to the same position stably even when the rotation angle is close to $\pi/2$ [equivalent to using a large learning rate according to Eq. (\ref{eq-angle})]. With a reasonable rotation angle such as $\pi/36$ or $\pi/18$, TSGO converges within 10 epochs. On the contrary, Adam suffers heavily from gradient vanishing and exploding problems. For the learning rate $\eta$ from $10^{-5}$ to $10^{-4}$, the training process converges to a higher value of the loss function, which indicates the gradient vanishing problem. The optimization becomes unstable when the learning rate is higher than $10^{-3}$, which indicates the gradient exploding problem.

\begin{figure}[htb]
	\includegraphics[width=1\linewidth]{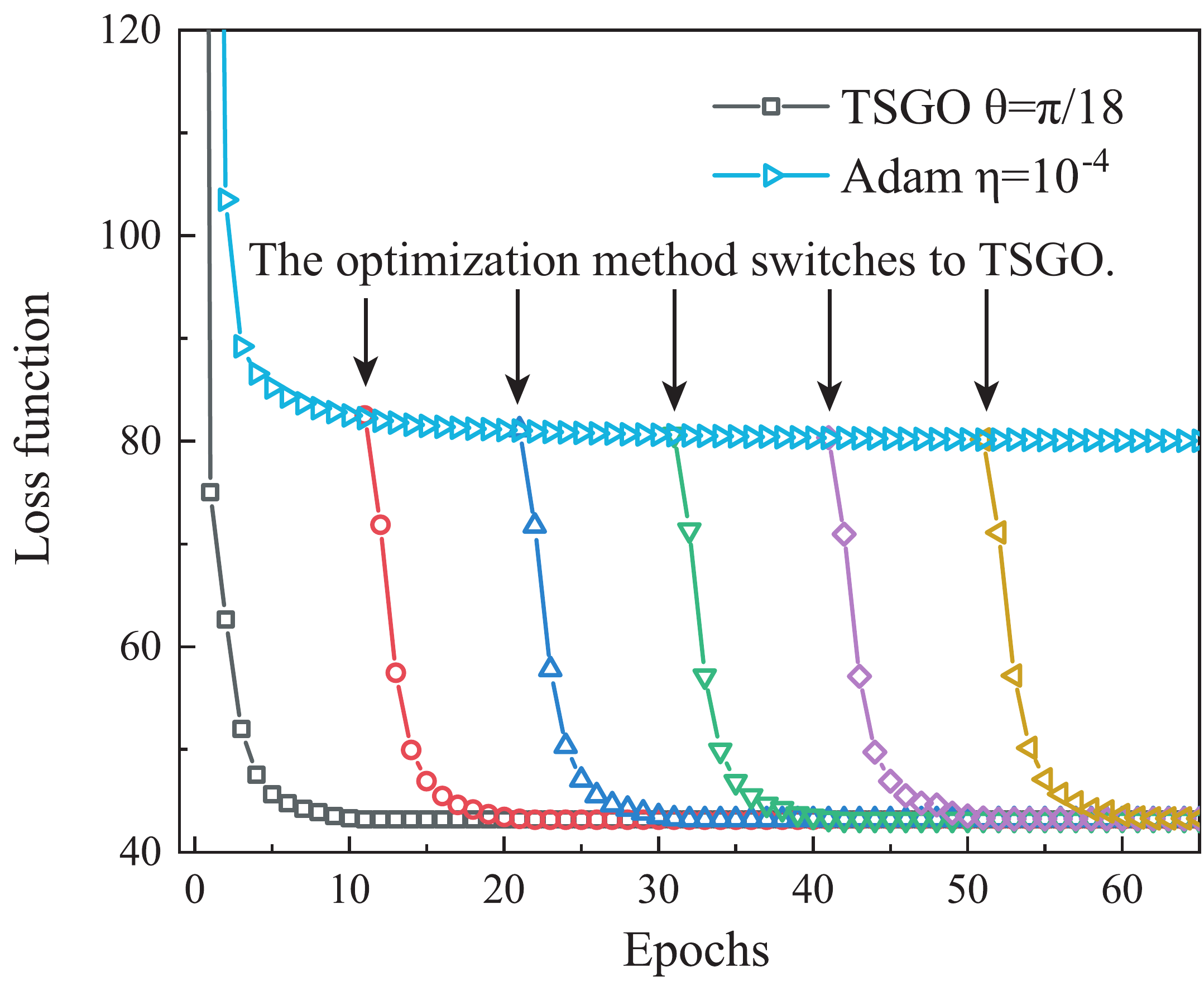}
	\caption{\label{retrain} Loss function versus epoch by TSGO and Adam. By switching the optimization method from Adam to TSGO, the loss function drops from around 80 to 40.}
	\end{figure}

To further verify that TSGO avoids the possible gradient vanishing problem, we firstly apply Adam and then switch to TSGO after certain epochs. Fig. \ref{retrain} shows that the loss function seems to converge by Adam, but immediately drops to a lower loss as soon as TSGO is implemented. Apparently, the state $|\psi \rangle $ optimized by Adam can still be corrected by TSGO.

The calculation of the gradient of deep NN involves the multiplications of a chain of matrices. Suppose that the gradient is calculated by repeatedly multiplying a matrix $M$ for $N$ times. The eigenvalue decomposition of $M$ is $M = U\Lambda {U^{ - 1}}$ with $U$ the transformation matrix. Then the gradient becomes $U{\Lambda ^N}{U^{ - 1}}$, where the eigenvalues are scaled as ${\Lambda^N}$. Therefore, any eigenvalues will either explode if they are greater than 1 or vanish if they are less than 1. MPS suffers the same difficulty, where the length of an MPS corresponds to the depth of an NN. If the MPS is (or close to be) normalized, the eigenvalues are in general smaller than 1, and one will mostly encounter gradient vanishing problems.

\begin{figure}[htb]
		\includegraphics[width=1\linewidth]{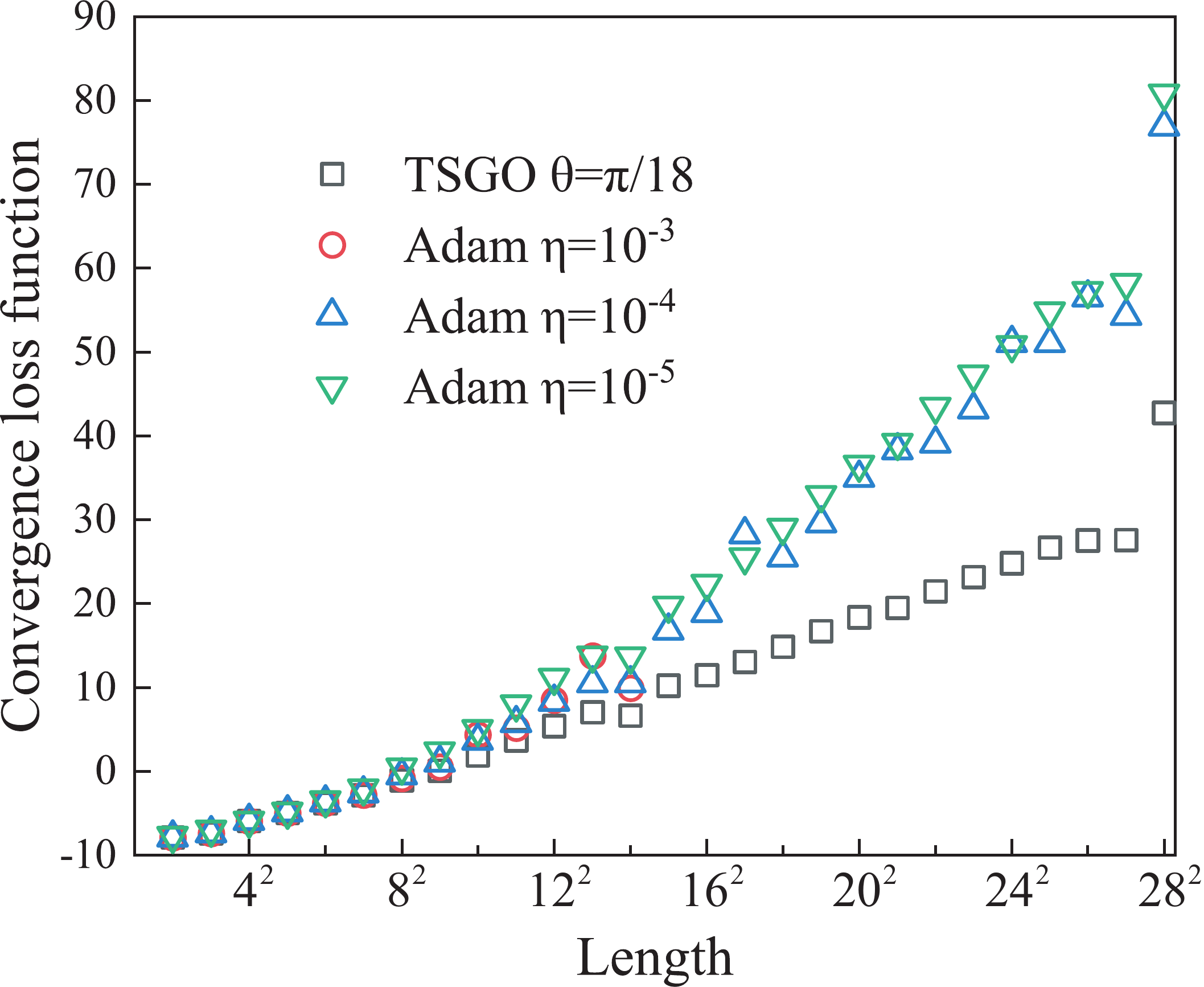}
		\caption{\label{compare_size} The convergent loss functions of the TSGO and Adam methods versus different lengths of the MPS's. Adam with $\eta  = {10^{ - 3}}$ is unstable when the length of MPS is larger than ${14^2}$.}
	\end{figure}

    To verify this, we give the convergent loss functions with different lengths $N$ of the MPS's (Fig .\ref{compare_size}).
Note that $N$ should be equal to the number of features. On MNIST, we control $N$ by resizing definitions of the images. For approximately $N<100$, there only exist small differences between the TSGO and Adam. However, for larger $N$'s, the BP algorithm with Adam is trapped to worse convergence. TSGO shows clearly better convergent loss functions.

    \textit{Conclusion and Discussion.---}We have introduced a tangent-space gradient optimization algorithm for probabilistic models to avoid gradient vanishing and exploding problems. The key point of TSGO is to restrict the gradient orthogonal to the parameter vector. The optimization of the model can be implemented by rotating parameter vector in Hilbert space. We show that TSGO brings a robust convergence that is independent of the learning rate and the depth of the model. In comparison, it is shown that the BP with Adam suffers the gradient vanishing and exploding problems for different learning rates and relatively large depth of the model.

    We shall note that the ideas of normalization are also used for avoiding the gradient vanishing/exploding problems for NN, such as weight normalization \cite{NIPS2016_6114}, batch normalization \cite{10.5555/3045118.3045167} and layer normalization \cite{ba2016layer}. With these methods, the predictions of the NN's are shown to be invariant under re-centering and re-scaling of the parameter vector. The normalization methods, therefore, have an implicit ``early stopping" effect and help to stabilize learning towards convergence \cite{ba2016layer}. TSGO gives more than the parameter invariance [see Eq. (\ref{eq-condition})] by revealing a explicit geometric relationship between the parameter vector and the gradient in the probabilistic model.

    We also note that TSGO in general cannot be implemented to update NN's, where we cannot guarantee Eq. (\ref{eq-condition}) in the presence of their high non-linearity. However, TSGO can in principle to be implemented in other probabilistic models, e.g., Boltzmann machines \cite{Boltzmann_Machines} or Bayesian networks \cite{jensen1996introduction}. We expect that TSGO would have crucial applications in developing new algorithms of machine learning.

\section*{Acknowledgments}

    This work is supported in part by the National Natural Science Foundation of China (11834014), the National Key R$\&$D Program of China (2018YFA0305800), and the Strategic Priority Research Program of the Chinese Academy of Sciences (XDB28000000). S.J.R. is also supported by Beijing Natural Science Foundation (Grant No. 1192005 and No. Z180013) and by the Academy for Multidisciplinary Studies, Capital Normal University.
\bibliography{ref}

\begin{thebibliography}{61}
\expandafter\ifx\csname natexlab\endcsname\relax\def\natexlab#1{#1}\fi
\expandafter\ifx\csname bibnamefont\endcsname\relax
  \def\bibnamefont#1{#1}\fi
\expandafter\ifx\csname bibfnamefont\endcsname\relax
  \def\bibfnamefont#1{#1}\fi
\expandafter\ifx\csname citenamefont\endcsname\relax
  \def\citenamefont#1{#1}\fi
\expandafter\ifx\csname url\endcsname\relax
  \def\url#1{\texttt{#1}}\fi
\expandafter\ifx\csname urlprefix\endcsname\relax\def\urlprefix{URL }\fi
\providecommand{\bibinfo}[2]{#2}
\providecommand{\eprint}[2][]{\url{#2}}

\bibitem[{\citenamefont{LeCun et~al.}(2015)\citenamefont{LeCun, Bengio, and
  Hinton}}]{lecun2015deep}
\bibinfo{author}{\bibfnamefont{Y.}~\bibnamefont{LeCun}},
  \bibinfo{author}{\bibfnamefont{Y.}~\bibnamefont{Bengio}}, \bibnamefont{and}
  \bibinfo{author}{\bibfnamefont{G.}~\bibnamefont{Hinton}},
  \bibinfo{journal}{nature} \textbf{\bibinfo{volume}{521}},
  \bibinfo{pages}{436} (\bibinfo{year}{2015}).

\bibitem[{\citenamefont{Deng and Yu}(2014)}]{SIG-039}
\bibinfo{author}{\bibfnamefont{L.}~\bibnamefont{Deng}} \bibnamefont{and}
  \bibinfo{author}{\bibfnamefont{D.}~\bibnamefont{Yu}},
  \bibinfo{journal}{Foundations and Trends® in Signal Processing}
  \textbf{\bibinfo{volume}{7}}, \bibinfo{pages}{197} (\bibinfo{year}{2014}),
  ISSN \bibinfo{issn}{1932-8346}.

\bibitem[{\citenamefont{Le et~al.}(2011)\citenamefont{Le, Ngiam, Coates,
  Lahiri, Prochnow, and Ng}}]{10.5555/3104482.3104516}
\bibinfo{author}{\bibfnamefont{Q.~V.} \bibnamefont{Le}},
  \bibinfo{author}{\bibfnamefont{J.}~\bibnamefont{Ngiam}},
  \bibinfo{author}{\bibfnamefont{A.}~\bibnamefont{Coates}},
  \bibinfo{author}{\bibfnamefont{A.}~\bibnamefont{Lahiri}},
  \bibinfo{author}{\bibfnamefont{B.}~\bibnamefont{Prochnow}}, \bibnamefont{and}
  \bibinfo{author}{\bibfnamefont{A.~Y.} \bibnamefont{Ng}}, in
  \emph{\bibinfo{booktitle}{Proceedings of the 28th International Conference on
  International Conference on Machine Learning}}
  (\bibinfo{publisher}{Omnipress}, \bibinfo{address}{Madison, WI, USA},
  \bibinfo{year}{2011}), ICML’11, p. \bibinfo{pages}{265–272}, ISBN
  \bibinfo{isbn}{9781450306195}.

\bibitem[{\citenamefont{Qian}(1999)}]{QIAN1999145}
\bibinfo{author}{\bibfnamefont{N.}~\bibnamefont{Qian}},
  \bibinfo{journal}{Neural Networks} \textbf{\bibinfo{volume}{12}},
  \bibinfo{pages}{145 } (\bibinfo{year}{1999}), ISSN \bibinfo{issn}{0893-6080}.

\bibitem[{\citenamefont{Kivinen and Warmuth}(1997)}]{KIVINEN19971}
\bibinfo{author}{\bibfnamefont{J.}~\bibnamefont{Kivinen}} \bibnamefont{and}
  \bibinfo{author}{\bibfnamefont{M.~K.} \bibnamefont{Warmuth}},
  \bibinfo{journal}{Information and Computation}
  \textbf{\bibinfo{volume}{132}}, \bibinfo{pages}{1 } (\bibinfo{year}{1997}),
  ISSN \bibinfo{issn}{0890-5401}.

\bibitem[{\citenamefont{Burges et~al.}(2005)\citenamefont{Burges, Shaked,
  Renshaw, Lazier, Deeds, Hamilton, and Hullender}}]{burges2005learning}
\bibinfo{author}{\bibfnamefont{C.}~\bibnamefont{Burges}},
  \bibinfo{author}{\bibfnamefont{T.}~\bibnamefont{Shaked}},
  \bibinfo{author}{\bibfnamefont{E.}~\bibnamefont{Renshaw}},
  \bibinfo{author}{\bibfnamefont{A.}~\bibnamefont{Lazier}},
  \bibinfo{author}{\bibfnamefont{M.}~\bibnamefont{Deeds}},
  \bibinfo{author}{\bibfnamefont{N.}~\bibnamefont{Hamilton}}, \bibnamefont{and}
  \bibinfo{author}{\bibfnamefont{G.~N.} \bibnamefont{Hullender}}, in
  \emph{\bibinfo{booktitle}{Proceedings of the 22nd International Conference on
  Machine learning (ICML-05)}} (\bibinfo{year}{2005}), pp.
  \bibinfo{pages}{89--96}.

\bibitem[{\citenamefont{Bottou}(2010)}]{10.1007/978-3-7908-2604-3_16}
\bibinfo{author}{\bibfnamefont{L.}~\bibnamefont{Bottou}}, in
  \emph{\bibinfo{booktitle}{Proceedings of COMPSTAT'2010}}, edited by
  \bibinfo{editor}{\bibfnamefont{Y.}~\bibnamefont{Lechevallier}}
  \bibnamefont{and} \bibinfo{editor}{\bibfnamefont{G.}~\bibnamefont{Saporta}}
  (\bibinfo{publisher}{Physica-Verlag HD}, \bibinfo{address}{Heidelberg},
  \bibinfo{year}{2010}), pp. \bibinfo{pages}{177--186}, ISBN
  \bibinfo{isbn}{978-3-7908-2604-3}.

\bibitem[{\citenamefont{Rojas}(1996)}]{Rojas1996}
\bibinfo{author}{\bibfnamefont{R.}~\bibnamefont{Rojas}},
  \emph{\bibinfo{title}{The Backpropagation Algorithm}}
  (\bibinfo{publisher}{Springer Berlin Heidelberg}, \bibinfo{address}{Berlin,
  Heidelberg}, \bibinfo{year}{1996}), pp. \bibinfo{pages}{149--182}, ISBN
  \bibinfo{isbn}{978-3-642-61068-4}.

\bibitem[{\citenamefont{Goodfellow et~al.}(2016)\citenamefont{Goodfellow,
  Bengio, and Courville}}]{Goodfellow-et-al-2016}
\bibinfo{author}{\bibfnamefont{I.}~\bibnamefont{Goodfellow}},
  \bibinfo{author}{\bibfnamefont{Y.}~\bibnamefont{Bengio}}, \bibnamefont{and}
  \bibinfo{author}{\bibfnamefont{A.}~\bibnamefont{Courville}},
  \emph{\bibinfo{title}{Deep Learning}} (\bibinfo{publisher}{MIT Press},
  \bibinfo{year}{2016}).

\bibitem[{\citenamefont{HECHT-NIELSEN}(1992)}]{HECHTNIELSEN199265}
\bibinfo{author}{\bibfnamefont{R.}~\bibnamefont{HECHT-NIELSEN}}, in
  \emph{\bibinfo{booktitle}{Neural Networks for Perception}}, edited by
  \bibinfo{editor}{\bibfnamefont{H.}~\bibnamefont{Wechsler}}
  (\bibinfo{publisher}{Academic Press}, \bibinfo{year}{1992}), pp.
  \bibinfo{pages}{65 -- 93}, ISBN \bibinfo{isbn}{978-0-12-741252-8}.

\bibitem[{\citenamefont{Ciresan et~al.}(2012)\citenamefont{Ciresan, Meier, and
  Schmidhuber}}]{DBLP:conf/cvpr/CiresanMS12}
\bibinfo{author}{\bibfnamefont{D.~C.} \bibnamefont{Ciresan}},
  \bibinfo{author}{\bibfnamefont{U.}~\bibnamefont{Meier}}, \bibnamefont{and}
  \bibinfo{author}{\bibfnamefont{J.}~\bibnamefont{Schmidhuber}}, in
  \emph{\bibinfo{booktitle}{2012 {IEEE} Conference on Computer Vision and
  Pattern Recognition, Providence, RI, USA, June 16-21, 2012}}
  (\bibinfo{year}{2012}), pp. \bibinfo{pages}{3642--3649}.

\bibitem[{\citenamefont{Krizhevsky et~al.}(2012)\citenamefont{Krizhevsky,
  Sutskever, and Hinton}}]{NIPS2012_4824}
\bibinfo{author}{\bibfnamefont{A.}~\bibnamefont{Krizhevsky}},
  \bibinfo{author}{\bibfnamefont{I.}~\bibnamefont{Sutskever}},
  \bibnamefont{and} \bibinfo{author}{\bibfnamefont{G.~E.}
  \bibnamefont{Hinton}}, in \emph{\bibinfo{booktitle}{Advances in Neural
  Information Processing Systems 25}}, edited by
  \bibinfo{editor}{\bibfnamefont{F.}~\bibnamefont{Pereira}},
  \bibinfo{editor}{\bibfnamefont{C.~J.~C.} \bibnamefont{Burges}},
  \bibinfo{editor}{\bibfnamefont{L.}~\bibnamefont{Bottou}}, \bibnamefont{and}
  \bibinfo{editor}{\bibfnamefont{K.~Q.} \bibnamefont{Weinberger}}
  (\bibinfo{publisher}{Curran Associates, Inc.}, \bibinfo{year}{2012}), pp.
  \bibinfo{pages}{1097--1105}.

\bibitem[{\citenamefont{Granter et~al.}(2017)\citenamefont{Granter, Beck, and
  Papke}}]{doi:10.5858/arpa.2016-0471-ED}
\bibinfo{author}{\bibfnamefont{S.~R.} \bibnamefont{Granter}},
  \bibinfo{author}{\bibfnamefont{A.~H.} \bibnamefont{Beck}}, \bibnamefont{and}
  \bibinfo{author}{\bibfnamefont{D.~J.} \bibnamefont{Papke}},
  \bibinfo{journal}{Archives of Pathology \& Laboratory Medicine}
  \textbf{\bibinfo{volume}{141}}, \bibinfo{pages}{619} (\bibinfo{year}{2017}),
  \bibinfo{note}{pMID: 28447900},
  \eprint{https://doi.org/10.5858/arpa.2016-0471-ED}.

\bibitem[{\citenamefont{Robbins and Monro}(1951)}]{10.2307/2236626}
\bibinfo{author}{\bibfnamefont{H.}~\bibnamefont{Robbins}} \bibnamefont{and}
  \bibinfo{author}{\bibfnamefont{S.}~\bibnamefont{Monro}},
  \bibinfo{journal}{The Annals of Mathematical Statistics}
  \textbf{\bibinfo{volume}{22}}, \bibinfo{pages}{400} (\bibinfo{year}{1951}).

\bibitem[{\citenamefont{Kushner and Yin}(2003)}]{kushner2003stochastic}
\bibinfo{author}{\bibfnamefont{H.}~\bibnamefont{Kushner}} \bibnamefont{and}
  \bibinfo{author}{\bibfnamefont{G.~G.} \bibnamefont{Yin}},
  \emph{\bibinfo{title}{Stochastic approximation and recursive algorithms and
  applications}}, vol.~\bibinfo{volume}{35} (\bibinfo{publisher}{Springer
  Science \& Business Media}, \bibinfo{year}{2003}).

\bibitem[{\citenamefont{Tieleman and Hinton}(2012)}]{tieleman2012lecture}
\bibinfo{author}{\bibfnamefont{T.}~\bibnamefont{Tieleman}} \bibnamefont{and}
  \bibinfo{author}{\bibfnamefont{G.}~\bibnamefont{Hinton}},
  \bibinfo{journal}{COURSERA: Neural networks for machine learning}
  \textbf{\bibinfo{volume}{4}}, \bibinfo{pages}{26} (\bibinfo{year}{2012}).

\bibitem[{\citenamefont{Zeiler}(2012)}]{zeiler2012adadelta}
\bibinfo{author}{\bibfnamefont{M.~D.} \bibnamefont{Zeiler}}
  (\bibinfo{year}{2012}), \eprint{arXiv:1212.5701}.

\bibitem[{\citenamefont{Kingma and Ba}(2015)}]{DBLP:journals/corr/KingmaB14}
\bibinfo{author}{\bibfnamefont{D.~P.} \bibnamefont{Kingma}} \bibnamefont{and}
  \bibinfo{author}{\bibfnamefont{J.}~\bibnamefont{Ba}}, in
  \emph{\bibinfo{booktitle}{3rd International Conference on Learning
  Representations, {ICLR} 2015, San Diego, CA, USA, May 7-9, 2015, Conference
  Track Proceedings}}, edited by
  \bibinfo{editor}{\bibfnamefont{Y.}~\bibnamefont{Bengio}} \bibnamefont{and}
  \bibinfo{editor}{\bibfnamefont{Y.}~\bibnamefont{LeCun}}
  (\bibinfo{year}{2015}).

\bibitem[{\citenamefont{Verstraete et~al.}(2008)\citenamefont{Verstraete, Murg,
  and Cirac}}]{verstraete2008matrix}
\bibinfo{author}{\bibfnamefont{F.}~\bibnamefont{Verstraete}},
  \bibinfo{author}{\bibfnamefont{V.}~\bibnamefont{Murg}}, \bibnamefont{and}
  \bibinfo{author}{\bibfnamefont{J.~I.} \bibnamefont{Cirac}},
  \bibinfo{journal}{Advances in Physics} \textbf{\bibinfo{volume}{57}},
  \bibinfo{pages}{143} (\bibinfo{year}{2008}).

\bibitem[{\citenamefont{Ran et~al.}(2020)\citenamefont{Ran, Tirrito, Peng,
  Chen, Su, and Lewenstein}}]{ran2017review}
\bibinfo{author}{\bibfnamefont{S.-J.} \bibnamefont{Ran}},
  \bibinfo{author}{\bibfnamefont{E.}~\bibnamefont{Tirrito}},
  \bibinfo{author}{\bibfnamefont{C.}~\bibnamefont{Peng}},
  \bibinfo{author}{\bibfnamefont{X.}~\bibnamefont{Chen}},
  \bibinfo{author}{\bibfnamefont{G.}~\bibnamefont{Su}}, \bibnamefont{and}
  \bibinfo{author}{\bibfnamefont{M.}~\bibnamefont{Lewenstein}},
  \emph{\bibinfo{title}{Tensor Network Contractions}}, vol.
  \bibinfo{volume}{964} of \emph{\bibinfo{series}{Lecture Notes in Physics}}
  (\bibinfo{publisher}{Springer International Publishing, Heidelberg},
  \bibinfo{year}{2020}), \bibinfo{edition}{1st} ed., ISBN
  \bibinfo{isbn}{978-3-030-34488-7}.

\bibitem[{\citenamefont{Evenbly and Vidal}(2011)}]{Evenbly2011}
\bibinfo{author}{\bibfnamefont{G.}~\bibnamefont{Evenbly}} \bibnamefont{and}
  \bibinfo{author}{\bibfnamefont{G.}~\bibnamefont{Vidal}},
  \bibinfo{journal}{Journal of Statistical Physics}
  \textbf{\bibinfo{volume}{145}}, \bibinfo{pages}{891} (\bibinfo{year}{2011}).

\bibitem[{\citenamefont{Bridgeman and Chubb}(2017)}]{bridgeman2017hand}
\bibinfo{author}{\bibfnamefont{J.~C.} \bibnamefont{Bridgeman}}
  \bibnamefont{and} \bibinfo{author}{\bibfnamefont{C.~T.} \bibnamefont{Chubb}},
  \bibinfo{journal}{Journal of Physics A: Mathematical and Theoretical}
  \textbf{\bibinfo{volume}{50}}, \bibinfo{pages}{223001}
  (\bibinfo{year}{2017}).

\bibitem[{\citenamefont{Schollwöck}(2011)}]{SCHOLLWOCK201196}
\bibinfo{author}{\bibfnamefont{U.}~\bibnamefont{Schollwöck}},
  \bibinfo{journal}{Annals of Physics} \textbf{\bibinfo{volume}{326}},
  \bibinfo{pages}{96 } (\bibinfo{year}{2011}), \bibinfo{note}{january 2011
  Special Issue}.

\bibitem[{\citenamefont{Cirac and Verstraete}(2009)}]{Cirac_2009}
\bibinfo{author}{\bibfnamefont{J.~I.} \bibnamefont{Cirac}} \bibnamefont{and}
  \bibinfo{author}{\bibfnamefont{F.}~\bibnamefont{Verstraete}},
  \bibinfo{journal}{Journal of Physics A: Mathematical and Theoretical}
  \textbf{\bibinfo{volume}{42}}, \bibinfo{pages}{504004}
  (\bibinfo{year}{2009}).

\bibitem[{\citenamefont{Orús}(2014)}]{ORUS2014117}
\bibinfo{author}{\bibfnamefont{R.}~\bibnamefont{Orús}},
  \bibinfo{journal}{Annals of Physics} \textbf{\bibinfo{volume}{349}},
  \bibinfo{pages}{117 } (\bibinfo{year}{2014}).

\bibitem[{\citenamefont{Cichocki et~al.}(2016)\citenamefont{Cichocki, Lee,
  Oseledets, Phan, Zhao, and Mandic}}]{MAL-059}
\bibinfo{author}{\bibfnamefont{A.}~\bibnamefont{Cichocki}},
  \bibinfo{author}{\bibfnamefont{N.}~\bibnamefont{Lee}},
  \bibinfo{author}{\bibfnamefont{I.}~\bibnamefont{Oseledets}},
  \bibinfo{author}{\bibfnamefont{A.-H.} \bibnamefont{Phan}},
  \bibinfo{author}{\bibfnamefont{Q.}~\bibnamefont{Zhao}}, \bibnamefont{and}
  \bibinfo{author}{\bibfnamefont{D.~P.} \bibnamefont{Mandic}},
  \bibinfo{journal}{Foundations and Trends® in Machine Learning}
  \textbf{\bibinfo{volume}{9}}, \bibinfo{pages}{249} (\bibinfo{year}{2016}).

\bibitem[{\citenamefont{Cichocki et~al.}(2017)\citenamefont{Cichocki, Phan,
  Zhao, Lee, Oseledets, Sugiyama, and Mandic}}]{MAL-067}
\bibinfo{author}{\bibfnamefont{A.}~\bibnamefont{Cichocki}},
  \bibinfo{author}{\bibfnamefont{A.-H.} \bibnamefont{Phan}},
  \bibinfo{author}{\bibfnamefont{Q.}~\bibnamefont{Zhao}},
  \bibinfo{author}{\bibfnamefont{N.}~\bibnamefont{Lee}},
  \bibinfo{author}{\bibfnamefont{I.}~\bibnamefont{Oseledets}},
  \bibinfo{author}{\bibfnamefont{M.}~\bibnamefont{Sugiyama}}, \bibnamefont{and}
  \bibinfo{author}{\bibfnamefont{D.~P.} \bibnamefont{Mandic}},
  \bibinfo{journal}{Foundations and Trends® in Machine Learning}
  \textbf{\bibinfo{volume}{9}}, \bibinfo{pages}{431} (\bibinfo{year}{2017}).

\bibitem[{\citenamefont{Biamonte et~al.}(2017)\citenamefont{Biamonte, Wittek,
  Pancotti, Rebentrost, Wiebe, and Lloyd}}]{biamonte2017quantum}
\bibinfo{author}{\bibfnamefont{J.}~\bibnamefont{Biamonte}},
  \bibinfo{author}{\bibfnamefont{P.}~\bibnamefont{Wittek}},
  \bibinfo{author}{\bibfnamefont{N.}~\bibnamefont{Pancotti}},
  \bibinfo{author}{\bibfnamefont{P.}~\bibnamefont{Rebentrost}},
  \bibinfo{author}{\bibfnamefont{N.}~\bibnamefont{Wiebe}}, \bibnamefont{and}
  \bibinfo{author}{\bibfnamefont{S.}~\bibnamefont{Lloyd}},
  \bibinfo{journal}{Nature} \textbf{\bibinfo{volume}{549}},
  \bibinfo{pages}{195} (\bibinfo{year}{2017}).

\bibitem[{\citenamefont{Huggins et~al.}(2019)\citenamefont{Huggins, Patil,
  Mitchell, Whaley, and Stoudenmire}}]{Huggins_2019}
\bibinfo{author}{\bibfnamefont{W.}~\bibnamefont{Huggins}},
  \bibinfo{author}{\bibfnamefont{P.}~\bibnamefont{Patil}},
  \bibinfo{author}{\bibfnamefont{B.}~\bibnamefont{Mitchell}},
  \bibinfo{author}{\bibfnamefont{K.~B.} \bibnamefont{Whaley}},
  \bibnamefont{and} \bibinfo{author}{\bibfnamefont{E.~M.}
  \bibnamefont{Stoudenmire}}, \bibinfo{journal}{Quantum Science and Technology}
  \textbf{\bibinfo{volume}{4}}, \bibinfo{pages}{024001} (\bibinfo{year}{2019}).

\bibitem[{\citenamefont{Stoudenmire and Schwab}(2016)}]{NIPS2016_6211}
\bibinfo{author}{\bibfnamefont{E.}~\bibnamefont{Stoudenmire}} \bibnamefont{and}
  \bibinfo{author}{\bibfnamefont{D.~J.} \bibnamefont{Schwab}}, in
  \emph{\bibinfo{booktitle}{Advances in Neural Information Processing Systems
  29}}, edited by \bibinfo{editor}{\bibfnamefont{D.~D.} \bibnamefont{Lee}},
  \bibinfo{editor}{\bibfnamefont{M.}~\bibnamefont{Sugiyama}},
  \bibinfo{editor}{\bibfnamefont{U.~V.} \bibnamefont{Luxburg}},
  \bibinfo{editor}{\bibfnamefont{I.}~\bibnamefont{Guyon}}, \bibnamefont{and}
  \bibinfo{editor}{\bibfnamefont{R.}~\bibnamefont{Garnett}}
  (\bibinfo{publisher}{Curran Associates, Inc.}, \bibinfo{year}{2016}), pp.
  \bibinfo{pages}{4799--4807}.

\bibitem[{\citenamefont{Glasser et~al.}(2018)\citenamefont{Glasser, Pancotti,
  and Cirac}}]{glasser2018supervised}
\bibinfo{author}{\bibfnamefont{I.}~\bibnamefont{Glasser}},
  \bibinfo{author}{\bibfnamefont{N.}~\bibnamefont{Pancotti}}, \bibnamefont{and}
  \bibinfo{author}{\bibfnamefont{J.~I.} \bibnamefont{Cirac}},
  \bibinfo{journal}{arXiv:1806.05964}  (\bibinfo{year}{2018}).

\bibitem[{\citenamefont{Chen et~al.}(2018)\citenamefont{Chen, Cheng, Xie, Wang,
  and Xiang}}]{PhysRevB.97.085104}
\bibinfo{author}{\bibfnamefont{J.}~\bibnamefont{Chen}},
  \bibinfo{author}{\bibfnamefont{S.}~\bibnamefont{Cheng}},
  \bibinfo{author}{\bibfnamefont{H.}~\bibnamefont{Xie}},
  \bibinfo{author}{\bibfnamefont{L.}~\bibnamefont{Wang}}, \bibnamefont{and}
  \bibinfo{author}{\bibfnamefont{T.}~\bibnamefont{Xiang}},
  \bibinfo{journal}{Phys. Rev. B} \textbf{\bibinfo{volume}{97}},
  \bibinfo{pages}{085104} (\bibinfo{year}{2018}).

\bibitem[{\citenamefont{Stoudenmire}(2018)}]{stoudenmire2018learning}
\bibinfo{author}{\bibfnamefont{E.~M.} \bibnamefont{Stoudenmire}},
  \bibinfo{journal}{Quantum Science and Technology}
  \textbf{\bibinfo{volume}{3}}, \bibinfo{pages}{034003} (\bibinfo{year}{2018}).

\bibitem[{\citenamefont{Han et~al.}(2018)\citenamefont{Han, Wang, Fan, Wang,
  and Zhang}}]{PhysRevX.8.031012}
\bibinfo{author}{\bibfnamefont{Z.-Y.} \bibnamefont{Han}},
  \bibinfo{author}{\bibfnamefont{J.}~\bibnamefont{Wang}},
  \bibinfo{author}{\bibfnamefont{H.}~\bibnamefont{Fan}},
  \bibinfo{author}{\bibfnamefont{L.}~\bibnamefont{Wang}}, \bibnamefont{and}
  \bibinfo{author}{\bibfnamefont{P.}~\bibnamefont{Zhang}},
  \bibinfo{journal}{Phys. Rev. X} \textbf{\bibinfo{volume}{8}},
  \bibinfo{pages}{031012} (\bibinfo{year}{2018}).

\bibitem[{\citenamefont{Liu et~al.}(2018)\citenamefont{Liu, Zhang, Lewenstein,
  and Ran}}]{liu2018learning}
\bibinfo{author}{\bibfnamefont{Y.}~\bibnamefont{Liu}},
  \bibinfo{author}{\bibfnamefont{X.}~\bibnamefont{Zhang}},
  \bibinfo{author}{\bibfnamefont{M.}~\bibnamefont{Lewenstein}},
  \bibnamefont{and} \bibinfo{author}{\bibfnamefont{S.-J.} \bibnamefont{Ran}},
  \bibinfo{journal}{arXiv:1803.09111}  (\bibinfo{year}{2018}).

\bibitem[{\citenamefont{Guo et~al.}(2018)\citenamefont{Guo, Jie, Lu, and
  Poletti}}]{PhysRevE.98.042114}
\bibinfo{author}{\bibfnamefont{C.}~\bibnamefont{Guo}},
  \bibinfo{author}{\bibfnamefont{Z.}~\bibnamefont{Jie}},
  \bibinfo{author}{\bibfnamefont{W.}~\bibnamefont{Lu}}, \bibnamefont{and}
  \bibinfo{author}{\bibfnamefont{D.}~\bibnamefont{Poletti}},
  \bibinfo{journal}{Phys. Rev. E} \textbf{\bibinfo{volume}{98}},
  \bibinfo{pages}{042114} (\bibinfo{year}{2018}).

\bibitem[{\citenamefont{Liu et~al.}(2019)\citenamefont{Liu, Ran, Wittek, Peng,
  Garc{\'{\i}}a, Su, and Lewenstein}}]{Liu_2019}
\bibinfo{author}{\bibfnamefont{D.}~\bibnamefont{Liu}},
  \bibinfo{author}{\bibfnamefont{S.-J.} \bibnamefont{Ran}},
  \bibinfo{author}{\bibfnamefont{P.}~\bibnamefont{Wittek}},
  \bibinfo{author}{\bibfnamefont{C.}~\bibnamefont{Peng}},
  \bibinfo{author}{\bibfnamefont{R.~B.} \bibnamefont{Garc{\'{\i}}a}},
  \bibinfo{author}{\bibfnamefont{G.}~\bibnamefont{Su}}, \bibnamefont{and}
  \bibinfo{author}{\bibfnamefont{M.}~\bibnamefont{Lewenstein}},
  \bibinfo{journal}{New Journal of Physics} \textbf{\bibinfo{volume}{21}},
  \bibinfo{pages}{073059} (\bibinfo{year}{2019}).

\bibitem[{\citenamefont{Cheng et~al.}(2019)\citenamefont{Cheng, Wang, Xiang,
  and Zhang}}]{PhysRevB.99.155131}
\bibinfo{author}{\bibfnamefont{S.}~\bibnamefont{Cheng}},
  \bibinfo{author}{\bibfnamefont{L.}~\bibnamefont{Wang}},
  \bibinfo{author}{\bibfnamefont{T.}~\bibnamefont{Xiang}}, \bibnamefont{and}
  \bibinfo{author}{\bibfnamefont{P.}~\bibnamefont{Zhang}},
  \bibinfo{journal}{Phys. Rev. B} \textbf{\bibinfo{volume}{99}},
  \bibinfo{pages}{155131} (\bibinfo{year}{2019}).

\bibitem[{\citenamefont{Pestun and Vlassopoulos}(2017)}]{pestun2017tensor}
\bibinfo{author}{\bibfnamefont{V.}~\bibnamefont{Pestun}} \bibnamefont{and}
  \bibinfo{author}{\bibfnamefont{Y.}~\bibnamefont{Vlassopoulos}}
  (\bibinfo{year}{2017}), \eprint{arXiv:1710.10248}.

\bibitem[{\citenamefont{Goodfellow et~al.}(2013)\citenamefont{Goodfellow,
  Mirza, Xiao, Courville, and Bengio}}]{goodfellow2013empirical}
\bibinfo{author}{\bibfnamefont{I.~J.} \bibnamefont{Goodfellow}},
  \bibinfo{author}{\bibfnamefont{M.}~\bibnamefont{Mirza}},
  \bibinfo{author}{\bibfnamefont{D.}~\bibnamefont{Xiao}},
  \bibinfo{author}{\bibfnamefont{A.}~\bibnamefont{Courville}},
  \bibnamefont{and} \bibinfo{author}{\bibfnamefont{Y.}~\bibnamefont{Bengio}}
  (\bibinfo{year}{2013}), \eprint{arXiv:1312.6211}.

\bibitem[{\citenamefont{Goodfellow et~al.}(2014)\citenamefont{Goodfellow,
  Pouget-Abadie, Mirza, Xu, Warde-Farley, Ozair, Courville, and
  Bengio}}]{NIPS2014_5423}
\bibinfo{author}{\bibfnamefont{I.}~\bibnamefont{Goodfellow}},
  \bibinfo{author}{\bibfnamefont{J.}~\bibnamefont{Pouget-Abadie}},
  \bibinfo{author}{\bibfnamefont{M.}~\bibnamefont{Mirza}},
  \bibinfo{author}{\bibfnamefont{B.}~\bibnamefont{Xu}},
  \bibinfo{author}{\bibfnamefont{D.}~\bibnamefont{Warde-Farley}},
  \bibinfo{author}{\bibfnamefont{S.}~\bibnamefont{Ozair}},
  \bibinfo{author}{\bibfnamefont{A.}~\bibnamefont{Courville}},
  \bibnamefont{and} \bibinfo{author}{\bibfnamefont{Y.}~\bibnamefont{Bengio}},
  in \emph{\bibinfo{booktitle}{Advances in Neural Information Processing
  Systems 27}}, edited by
  \bibinfo{editor}{\bibfnamefont{Z.}~\bibnamefont{Ghahramani}},
  \bibinfo{editor}{\bibfnamefont{M.}~\bibnamefont{Welling}},
  \bibinfo{editor}{\bibfnamefont{C.}~\bibnamefont{Cortes}},
  \bibinfo{editor}{\bibfnamefont{N.~D.} \bibnamefont{Lawrence}},
  \bibnamefont{and} \bibinfo{editor}{\bibfnamefont{K.~Q.}
  \bibnamefont{Weinberger}} (\bibinfo{publisher}{Curran Associates, Inc.},
  \bibinfo{year}{2014}), pp. \bibinfo{pages}{2672--2680}.

\bibitem[{\citenamefont{Tang et~al.}(2018)\citenamefont{Tang, Xiao, Li, and
  Wang}}]{TANG2018125}
\bibinfo{author}{\bibfnamefont{H.}~\bibnamefont{Tang}},
  \bibinfo{author}{\bibfnamefont{B.}~\bibnamefont{Xiao}},
  \bibinfo{author}{\bibfnamefont{W.}~\bibnamefont{Li}}, \bibnamefont{and}
  \bibinfo{author}{\bibfnamefont{G.}~\bibnamefont{Wang}},
  \bibinfo{journal}{Information Sciences} \textbf{\bibinfo{volume}{433-434}},
  \bibinfo{pages}{125 } (\bibinfo{year}{2018}).

\bibitem[{\citenamefont{Cheng et~al.}(2018)\citenamefont{Cheng, Chen, and
  Wang}}]{e20080583}
\bibinfo{author}{\bibfnamefont{S.}~\bibnamefont{Cheng}},
  \bibinfo{author}{\bibfnamefont{J.}~\bibnamefont{Chen}}, \bibnamefont{and}
  \bibinfo{author}{\bibfnamefont{L.}~\bibnamefont{Wang}},
  \bibinfo{journal}{Entropy} \textbf{\bibinfo{volume}{20}}
  (\bibinfo{year}{2018}).

\bibitem[{\citenamefont{P{\'{e}}rez{-}Garc{\'{\i}}a
  et~al.}(2007)\citenamefont{P{\'{e}}rez{-}Garc{\'{\i}}a, Verstraete, Wolf, and
  Cirac}}]{DBLP:journals/qic/Perez-GarciaVWC07}
\bibinfo{author}{\bibfnamefont{D.}~\bibnamefont{P{\'{e}}rez{-}Garc{\'{\i}}a}},
  \bibinfo{author}{\bibfnamefont{F.}~\bibnamefont{Verstraete}},
  \bibinfo{author}{\bibfnamefont{M.~M.} \bibnamefont{Wolf}}, \bibnamefont{and}
  \bibinfo{author}{\bibfnamefont{J.~I.} \bibnamefont{Cirac}},
  \bibinfo{journal}{Quantum Information {\&} Computation}
  \textbf{\bibinfo{volume}{7}}, \bibinfo{pages}{401} (\bibinfo{year}{2007}).

\bibitem[{\citenamefont{Shi et~al.}(2006)\citenamefont{Shi, Duan, and
  Vidal}}]{PhysRevA.74.022320}
\bibinfo{author}{\bibfnamefont{Y.-Y.} \bibnamefont{Shi}},
  \bibinfo{author}{\bibfnamefont{L.-M.} \bibnamefont{Duan}}, \bibnamefont{and}
  \bibinfo{author}{\bibfnamefont{G.}~\bibnamefont{Vidal}},
  \bibinfo{journal}{Phys. Rev. A} \textbf{\bibinfo{volume}{74}},
  \bibinfo{pages}{022320} (\bibinfo{year}{2006}).

\bibitem[{\citenamefont{Cincio et~al.}(2008)\citenamefont{Cincio, Dziarmaga,
  and Rams}}]{PhysRevLett.100.240603}
\bibinfo{author}{\bibfnamefont{L.}~\bibnamefont{Cincio}},
  \bibinfo{author}{\bibfnamefont{J.}~\bibnamefont{Dziarmaga}},
  \bibnamefont{and} \bibinfo{author}{\bibfnamefont{M.~M.} \bibnamefont{Rams}},
  \bibinfo{journal}{Phys. Rev. Lett.} \textbf{\bibinfo{volume}{100}},
  \bibinfo{pages}{240603} (\bibinfo{year}{2008}).

\bibitem[{\citenamefont{Born}(1926)}]{born1926quantum}
\bibinfo{author}{\bibfnamefont{M.}~\bibnamefont{Born}}, \bibinfo{journal}{Zeit
  fur Phys} \textbf{\bibinfo{volume}{38}}, \bibinfo{pages}{803}
  (\bibinfo{year}{1926}).

\bibitem[{\citenamefont{BALLENTINE}(1970)}]{RevModPhys.42.358}
\bibinfo{author}{\bibfnamefont{L.~E.} \bibnamefont{BALLENTINE}},
  \bibinfo{journal}{Rev. Mod. Phys.} \textbf{\bibinfo{volume}{42}},
  \bibinfo{pages}{358} (\bibinfo{year}{1970}).

\bibitem[{\citenamefont{Kullback and Leibler}(1951)}]{10.2307/2236703}
\bibinfo{author}{\bibfnamefont{S.}~\bibnamefont{Kullback}} \bibnamefont{and}
  \bibinfo{author}{\bibfnamefont{R.~A.} \bibnamefont{Leibler}},
  \bibinfo{journal}{The Annals of Mathematical Statistics}
  \textbf{\bibinfo{volume}{22}}, \bibinfo{pages}{79} (\bibinfo{year}{1951}).

\bibitem[{\citenamefont{Oseledets}(2011)}]{doi:10.1137/090752286}
\bibinfo{author}{\bibfnamefont{I.}~\bibnamefont{Oseledets}},
  \bibinfo{journal}{SIAM Journal on Scientific Computing}
  \textbf{\bibinfo{volume}{33}}, \bibinfo{pages}{2295} (\bibinfo{year}{2011}),
  \eprint{https://doi.org/10.1137/090752286}.

\bibitem[{\citenamefont{Tagliacozzo et~al.}(2009)\citenamefont{Tagliacozzo,
  Evenbly, and Vidal}}]{PhysRevB.80.235127}
\bibinfo{author}{\bibfnamefont{L.}~\bibnamefont{Tagliacozzo}},
  \bibinfo{author}{\bibfnamefont{G.}~\bibnamefont{Evenbly}}, \bibnamefont{and}
  \bibinfo{author}{\bibfnamefont{G.}~\bibnamefont{Vidal}},
  \bibinfo{journal}{Phys. Rev. B} \textbf{\bibinfo{volume}{80}},
  \bibinfo{pages}{235127} (\bibinfo{year}{2009}).

\bibitem[{\citenamefont{Verstraete and
  Cirac}(2004)}]{verstraete2004renormalization}
\bibinfo{author}{\bibfnamefont{F.}~\bibnamefont{Verstraete}} \bibnamefont{and}
  \bibinfo{author}{\bibfnamefont{J.~I.} \bibnamefont{Cirac}}
  (\bibinfo{year}{2004}), \eprint{cond-mat/0407066}.

\bibitem[{\citenamefont{Jordan et~al.}(2008)\citenamefont{Jordan, Or\'us,
  Vidal, Verstraete, and Cirac}}]{PhysRevLett.101.250602}
\bibinfo{author}{\bibfnamefont{J.}~\bibnamefont{Jordan}},
  \bibinfo{author}{\bibfnamefont{R.}~\bibnamefont{Or\'us}},
  \bibinfo{author}{\bibfnamefont{G.}~\bibnamefont{Vidal}},
  \bibinfo{author}{\bibfnamefont{F.}~\bibnamefont{Verstraete}},
  \bibnamefont{and} \bibinfo{author}{\bibfnamefont{J.~I.} \bibnamefont{Cirac}},
  \bibinfo{journal}{Phys. Rev. Lett.} \textbf{\bibinfo{volume}{101}},
  \bibinfo{pages}{250602} (\bibinfo{year}{2008}).

\bibitem[{\citenamefont{Evenbly}(2018)}]{PhysRevB.98.085155}
\bibinfo{author}{\bibfnamefont{G.}~\bibnamefont{Evenbly}},
  \bibinfo{journal}{Phys. Rev. B} \textbf{\bibinfo{volume}{98}},
  \bibinfo{pages}{085155} (\bibinfo{year}{2018}).

\bibitem[{\citenamefont{Haghshenas et~al.}(2019)\citenamefont{Haghshenas,
  O'Rourke, and Chan}}]{haghshenas2019canonicalization}
\bibinfo{author}{\bibfnamefont{R.}~\bibnamefont{Haghshenas}},
  \bibinfo{author}{\bibfnamefont{M.~J.} \bibnamefont{O'Rourke}},
  \bibnamefont{and} \bibinfo{author}{\bibfnamefont{G.~K.} \bibnamefont{Chan}}
  (\bibinfo{year}{2019}), \eprint{arXiv:1903.03843}.

\bibitem[{\citenamefont{{Deng}}(2012)}]{6296535}
\bibinfo{author}{\bibfnamefont{L.}~\bibnamefont{{Deng}}},
  \bibinfo{journal}{IEEE Signal Processing Magazine}
  \textbf{\bibinfo{volume}{29}}, \bibinfo{pages}{141} (\bibinfo{year}{2012}).

\bibitem[{\citenamefont{Salimans and Kingma}(2016)}]{NIPS2016_6114}
\bibinfo{author}{\bibfnamefont{T.}~\bibnamefont{Salimans}} \bibnamefont{and}
  \bibinfo{author}{\bibfnamefont{D.~P.} \bibnamefont{Kingma}}, in
  \emph{\bibinfo{booktitle}{Advances in Neural Information Processing Systems
  29}}, edited by \bibinfo{editor}{\bibfnamefont{D.~D.} \bibnamefont{Lee}},
  \bibinfo{editor}{\bibfnamefont{M.}~\bibnamefont{Sugiyama}},
  \bibinfo{editor}{\bibfnamefont{U.~V.} \bibnamefont{Luxburg}},
  \bibinfo{editor}{\bibfnamefont{I.}~\bibnamefont{Guyon}}, \bibnamefont{and}
  \bibinfo{editor}{\bibfnamefont{R.}~\bibnamefont{Garnett}}
  (\bibinfo{publisher}{Curran Associates, Inc.}, \bibinfo{year}{2016}), pp.
  \bibinfo{pages}{901--909}.

\bibitem[{\citenamefont{Ioffe and Szegedy}(2015)}]{10.5555/3045118.3045167}
\bibinfo{author}{\bibfnamefont{S.}~\bibnamefont{Ioffe}} \bibnamefont{and}
  \bibinfo{author}{\bibfnamefont{C.}~\bibnamefont{Szegedy}}, in
  \emph{\bibinfo{booktitle}{Proceedings of the 32nd International Conference on
  International Conference on Machine Learning - Volume 37}}
  (\bibinfo{publisher}{JMLR.org}, \bibinfo{year}{2015}), ICML’15, p.
  \bibinfo{pages}{448–456}.

\bibitem[{\citenamefont{Ba et~al.}(2016)\citenamefont{Ba, Kiros, and
  Hinton}}]{ba2016layer}
\bibinfo{author}{\bibfnamefont{J.~L.} \bibnamefont{Ba}},
  \bibinfo{author}{\bibfnamefont{J.~R.} \bibnamefont{Kiros}}, \bibnamefont{and}
  \bibinfo{author}{\bibfnamefont{G.~E.} \bibnamefont{Hinton}}
  (\bibinfo{year}{2016}), \eprint{arXiv:1607.06450}.

\bibitem[{\citenamefont{Ackley et~al.}(1985)\citenamefont{Ackley, Hinton, and
  Sejnowski}}]{Boltzmann_Machines}
\bibinfo{author}{\bibfnamefont{D.~H.} \bibnamefont{Ackley}},
  \bibinfo{author}{\bibfnamefont{G.~E.} \bibnamefont{Hinton}},
  \bibnamefont{and} \bibinfo{author}{\bibfnamefont{T.~J.}
  \bibnamefont{Sejnowski}}, \bibinfo{journal}{Cognitive Science}
  \textbf{\bibinfo{volume}{9}}, \bibinfo{pages}{147} (\bibinfo{year}{1985}).

\bibitem[{\citenamefont{Jensen et~al.}(1996)}]{jensen1996introduction}
\bibinfo{author}{\bibfnamefont{F.~V.} \bibnamefont{Jensen}}
  \bibnamefont{et~al.}, \emph{\bibinfo{title}{An introduction to Bayesian
  networks}}, vol. \bibinfo{volume}{210} (\bibinfo{publisher}{UCL press
  London}, \bibinfo{year}{1996}).

\end{thebibliography}

\end{document}